\documentclass[]{spie}  %>>> use for US letter paper
\usepackage[]{graphicx}
\usepackage[]{amsmath}
\usepackage[]{amssymb}
\usepackage[]{caption}
\usepackage[]{algorithm2e}
\usepackage[]{subfig} 
\usepackage[]{multirow}
\usepackage[]{array}

\DeclareMathOperator*{\argmin}{arg\,min}
\DeclareMathOperator*{\dvg}{\operatorname{div}}

\title{A Continuous Max-Flow Approach to Multi-Labeling Problems under Arbitrary Region Regularization}

\author{John S.H. Baxter\supit{a,b}, Martin Rajchl\supit{a,b},
	Jing Yuan\supit{a,b}, and Terry M. Peters\supit{a,b}
\skiplinehalf
\supit{a}Robarts Research Institute, London, Ontario, Canada; \\
\supit{b}Western University, London, Ontario, Canada
}

\authorinfo{Send correspondence to J.S.H.B.: E-mail: jbaxter@robarts.ca}
\begin{document} 
  \maketitle 

%%%%%%%%%%%%%%%%%%%%%%%%%%%%%%%%%%%%%%%%%%%%%%%%%%%%%%%%%%%%% 
\begin{abstract}
The incorporation of region regularization into max-flow segmentation has traditionally focused on ordering and part-whole relationships. A side effect of the development of such models is that it constrained regularization only to those cases, rather than allowing for arbitrary region regularization. Directed Acyclic Graphical Max-Flow (DAGMF) segmentation overcomes these limitations by allowing for the algorithm designer to specify an arbitrary directed acyclic graph to structure a max-flow segmentation. This allows for individual `parts' to be a member of multiple distinct `wholes.'
\end{abstract}

%>>>> Include a list of keywords after the abstract 

%%%%%%%%%%%%%%%%%%%%%%%%%%%%%%%%%%%%%%%%%%%%%%%%%%%%%%%%%%%%%
\section{INTRODUCTION}
\label{sec:intro}
The incorporation of anatomical knowledge into multi-region medical image segmentation has been the subject of countless articles. Recently, research into the specification and incorporation of anatomy-agnostic knowledge structures have been undertaken with varying levels of success. For example, regularization has been used initially to encourage segmentation contiguity or adherence to image edge features, most notably popularized by the graph cuts segmentation approach \cite{boykov_interactive_2001,boykov_fast_2001,boykov_experimental_2004}. This framework represented the image as a finite lattice, incorporating regularization in a globally optimal manner, through the use of edge weighting and minimum cost cuts. Such an approach had direct analogues in Markov Random Field theory \cite{boykov_markov_1998,li_markov_2009} allowing for guided development of cost terms using standard probability theory. Such work has been extended from the discrete domain to a continuous domain without the loss of global optimality \cite{yuan_study_2010} mitigating the effects of differing neighbourhood connectivity and associated metrification artifacts.

Regularization was then extended to incorporate or encourage spatial grouping relationships, originally in the form of a full ordering using the discrete Ishikawa model \cite{ishikawa_exact_2003} and its continuous counterpart \cite{bae_fast_2011}. However, these models enforce that a full ordering be defined \textit{a priori} for the segmentation problem, and were not applicable to segmentation problems outside of that scope. This scope has been extended recently by Delong et al. \cite{delong2009globally} for the discrete case and Baxter et al. \cite{baxterGHMF2014} for the continuous, but maintained constraints on which regularization configurations could be specified.

The motivation behind this work is to extend previous general models to eliminate the constraints on what part/whole or ordering relationships can be defined in the continuous case. Thus, an algorithm for solving continuous max-flow/min-cut segmentation problems under a directed acyclic graph over-architecture is developed, along with a framework for expressing arbitrary super-object regularization using said algorithm.

\section{Contributions}
As with previous work in extensible max-flow segmentation models, this algorithm displays a high degree of inherent parallelism allowing for acceleration through general purpose graphic processing unit (GPGPU) computation, as well as additional over-arching concurrency allowing for additional threading and scheduling to improve performance and multi-card use.

\newpage
\section{Directed Acyclic Graphical Model and Previous Work}
\subsection{Previous Work}
Work by Yuan et al. \cite{yuan_study_2010} addressed both the continuous binary min-cut problem and the convex relaxed Potts model:
\begin{gather*}
\begin{array}{ccc}
E(u) = \int\limits_{\Omega}(D_s(x)u(x) + D_t(x)(1-u(x))+\alpha S(x)|\nabla u(x)|)dx & \, \, &
E(u) = \sum\limits_{\forall L} \int\limits_{\Omega}(D_L(x)u_L(x)+\alpha S(x)|\nabla u_L(x)|)dx \\
\text{ s.t. } u(x) \in \{ 0, 1 \} && 
\text{ s.t. } u_L(x) \geq 0 \mbox{ and } \sum\limits_{\forall L}u_L(x) = 1 \mbox{ .}
\end{array}
\end{gather*}
This work was further extended by Bae et al. \cite{bae_fast_2011} to the continuous Ishikawa model:
\begin{gather*}
E(u) = \sum\limits_{L=0}^N \int\limits_{\Omega}(D_L(x)u_L(x)+\alpha S(x)|\nabla u_L(x)|)dx \\
\mbox{ s.t. } u_L(x) \in \{ 0,1 \} \mbox{ and } u_{L+1}(x) \leq u_L(x) \text{ ,}
\end{gather*}
using similar variational methods but employed a tiered continuous graph analogous to that used by Ishikawa \cite{ishikawa_exact_2003} in the discrete case, that is, with finite capacities on intermediate flows between labels.

Models with limited hierarchical constraints such as that used by Rajchl et al.\cite{rajchl_interactive_2014} for myocardial scar segmentation have been posed. They have since been generalized by Baxter et al. \cite{baxterGHMF2014} in the form:
\begin{gather*}
E(u) = \sum\limits_{\forall L} \int_\Omega \left( D_L(x)u_L(x)+S_{L}(x)|\nabla u_L(x)| \right) dx \\
\mbox{ s.t. } \forall L (u_L(x) \geq 0) \text{ and } \forall L \left(u_L(x) = \sum\limits_{ L' \in L.C }u_{L'}(x) \right)  \text{ and }  u_S(x)=1
\end{gather*}

These techniques both used a continuous max-flow model with augmented Lagrangian multipliers from which efficient solution algorithms could be constructed. Apart from regularization structure, constraints such as star-shaped constraints on the various labels\cite{yuan_efficient_2012}, as well as volume preserving\cite{}, and inter-image consistency\cite{} has been incorporated.

\subsection{Directed Acyclic Graphical Model}
\begin{figure}[h]
\centering
\includegraphics[height=1.893in]{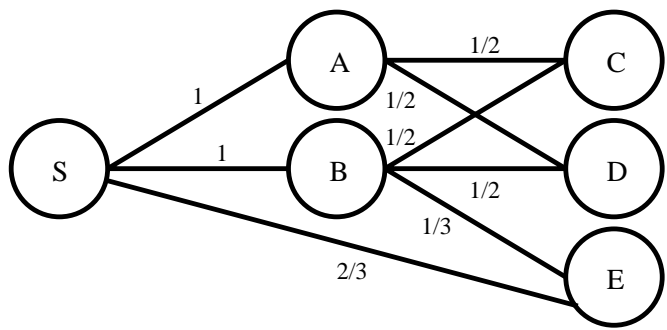}
\caption{Directed Acyclic Graph for Label Representation}
\label{fig:dag}
\end{figure}

Directed Acyclic Graphical Max-Flow (DAGMF) segmentation relies heavily on the concept of a rooted directed acyclic graph (DAG) with weighted edges for label representation. Rooted directed acyclic graphs extend the notion of part/whole or parent/child relationships used in previous models, specifically hierarchical models\cite{rajchl_interactive_2014,baxterGHMF2014}. For now, a discussion of interpretation and whole/part relationships will be postponed until Section \ref{sec:arbitraryReg}. Figure \ref{fig:dag} provides an example of this structure.

Throughout this paper, we will refer to parent and child operators, $.P$ and $.C$ respectively. These reflect the direction of flow (always from parent to children) and multiplier accumulation (always from child to parents). In addition, edges in this graph, and corresponding parent/child operators can have multiplicity, that is, multiple edges can exist between two vertices indicating some rational weighting of relative parts. The parent-child operators for the graph in Figure \ref{fig:dag} are:
\begin{equation*}
\begin{array}{lll}
S.P = \emptyset && S.C = \{ A, B, E\}\\
A.P = \{ S \} && A.C = \{ C, D\}\\
B.P = \{ S \} && B.C = \{ C, D, E \}\\
C.P = \{ A, B \} && C.C = \emptyset \\
D.P = \{ A, B \} && D.C = \emptyset \\
E.P = \{ A, B, S \} & \, \, & E.C = \emptyset \\
\end{array}
\end{equation*}
As with the general hierarchical formulation \cite{baxterGHMF2014}, these operators have consistency requirements. That is, they must form a rooted DAG with the properties:
\begin{itemize}\itemsep1pt \parskip0pt \parsep0pt
\item there exists only one label, $S$, has no parent ($S.P = \emptyset$), 
\item the parent/child relationship is preserved ($A\in B.P \leftrightarrow B \in A.C$),
\item there are no cycles (no label can be a child, grandchild, great-grandchild, etc... of itself) and that the graph is connected. Note that if this property holds for the child operator, it must also hold for the parent operator due to the above properties, and
\item the edge weights are non-negative and normalized for each child, that is $\forall L \, \sum_{L' \in L.C}w_{(L',L)} = 1$.
\end{itemize}

As with the Potts model \cite{potts_generalized_1952} we would like to define the set of end-labels, $\mathbb{L}$, as a partition of the image, that is:
\begin{equation}
\bigcup_{L \in \mathbb{L}} \Omega_L = \Omega \text{ and } \forall L_1,L_2 \in \mathbb{L} \, (L_1 \neq L_2 \implies L_1 \cap L_2 = \emptyset) \text{ .}
\end{equation}
Unlike those in the Ishikawa \cite{ishikawa_exact_2003} model and generalized hierarchical max-flow model \cite{baxterGHMF2014}, the intermediate labels do not immediately lend themselves to a set-theoretic interpretation since they are not constrained to be the union of end-labels. (Although, this case will be explicitly explored in Section \ref{sec:arbitraryReg} with corresponding DAG structure.) Nevertheless, the labeling function, $u_L(x) \in [0,1]$  has the following properties:\cite{yuan_study_2010, yuan_continuous_2010, bae_global_2011}
\begin{equation}
u_L(x) = \left\{ \begin{array}{ll}
 1, & x \in \Omega_L \\ 
 0, & x \not\in \Omega_L \\
 \end{array} \right.
\end{equation}
\begin{equation}
\int_{\delta \Omega_L}S_L(x)dx = \int_\Omega S_L(x) |\nabla u_L(x)| dx
\end{equation}
for labels in $\mathbb{L}$. In the case where the DAG forms a hierarchy, the same constraints apply as in the generalized hierarchical model. Otherwise, the non-unit multiplicands render the union operator meaningless. In terms of labeling function, the intermediate nodes are defined as:
\begin{equation}
u_L(x) = \sum\limits_{ L' \in L.C } w_{(L,L')} u_{L'}(x) \text{ .}
\end{equation}

These yield the convex relaxed generalized hierarchical model:
\begin{equation}
\begin{aligned}
& \underset{u}{\text{minimize}}
& & E(u) = \sum\limits_{\forall L} \int_\Omega \left( D_L(x)u_L(x)+S_{L}(x)|\nabla u_L(x)| \right) dx \\
& \text{subject to} & & \forall L (u_L(x) \geq 0) \\
& & &  \forall L \left(u_L(x) = \sum\limits_{ L' \in L.C }w_{(L,L')} u_{L'}(x)\right) \\
& & &   u_S(x)=1 \text{ .}
\end{aligned}
\label{dagmf_form}
\end{equation}
These formulations can be solved with global optimality for probabilistic labels, which will be demonstrated by this paper. However, it is only an approximation algorithm for models under integrality constraints on the end-labels, as demonstrated by the NP-hardness of the Potts model, even over finite lattices\cite{kolmogorov_computing_2001}.

As with previous approaches, we only consider the case where the end-labels have non-zero data terms. To prove that this does not limit the applicability of the method, a similar linear-time data term pushdown proof can be constructed along the lines presented by Baxter at al.\cite{baxterGHMF2014}

\section{Continuous Max-Flow Model}
\subsection{Primal Formulation}
The modeling approach is derived from those presented by Yuan et al.\cite{yuan_study_2010, yuan_continuous_2010}] and follows the same format, using duality through an augmented Langrangian formulation. The primal model represents network flow maximization through a large graph with only the sink flows constrained. The dual of this formulation is the DAGMF equation \eqref{dagmf_form} as we shall prove in this section. We can write the primal model as:
\begin{equation}
\label{primal}
\begin{aligned}
\underset{p,q} \max & \int_\Omega p_S(x) dx
\end{aligned}
\end{equation}
subject to the constraints
\begin{equation}
\begin{aligned}
p_L(x) & \leq D_L(x), \text{ where } L.C = \emptyset \\
|q_L(x)| & \leq \alpha_L S_L(x) \text{ } L \neq S \\
0 & =\operatorname{div} q_L(x) + p_L(x) - \sum_{L' \in L.P}w_{(L',L)}p_{L'}(x) \text{ .}
\end{aligned}
\end{equation}

This is equivalent to a multi-flow problem over a large graph constructed from the image dimensions and the provided directed acyclic graph as overall architecture. Other than constraints put on the magnitude of the spatial flows, and the capacity of the sink flows ($p_L(x)$ where $L.C = \emptyset$), the system is assumed to have infinite capacity. This is a strict generalization of the hierarchical formed explored by Baxter et al.\cite{baxterGHMF2014} considering hierarchies to be a specific class of rooted DAG.

\subsection{Primal-Dual Formulation}
The primal model can be converted to a primal-dual model through the use of Lagrangian multipliers over the flow conservation constraint $G_L(x) = \operatorname{div} q_L(x) + p_L(x) - \sum_{L' \in L.P}w_{(L',L)}p_{L'}(x) = 0$ yielding the Lagrangian:
\begin{equation}
\label{lagrangian}
\begin{aligned}
\underset{u} \min \underset{p,q} \max & \left( \int_\Omega p_S(x) dx + \sum_{\forall L \neq S} \int_\Omega u_L(x) G_L(x) dx \right) \\
p_L(x) & \leq D_L(x), \text{ where } L.C = \emptyset \\
|q_L(x)| & \leq \alpha_L S_L(x) \text{ } L \neq S \text{ .}
\end{aligned}
\end{equation}

First, we must ensure that equation \eqref{lagrangian} is convex with respect to $u$, considering $p,q$ to be fixed, and concave with respect to $p,q$ with $u$ fixed, as to meet the requirements of the minimax theorem. \cite{ekeland_convex_1976} Considering $p,q$ as fixed, $G$ is obviously fixed as well, implying that equation \eqref{lagrangian} is linear over $u$ and therefore convex. It should also be noted that $G$ is a linear function of $p,q$, meaning that \eqref{lagrangian} is again linear and therefore concave with respect to $p,q$ This implies the existence of a saddle point and the equivalence of the formulation regardless of the order of the prefix max and min operators under the minimax theorem. \cite{ekeland_convex_1976}

\newpage
\subsection{Dual Formulation}
As implied in the previous section, we can find the saddle point through the optimization of the sink-flows, $p_L$, working bottom-up and the spatial flows within each label. For the sake of simplicity, we refer to an ordering $\mathbb{O}$ generated by topologically sorting the graph $G = \left( \mathbb{V}\cup\{ S \}, \{ (L_1,L_2) | L_2 \in L_1.P \} \right)$. We proceed through the graph in that order. Starting with any label, $L$, such that $L.C = \emptyset$ we can isolate $p_L$ in \eqref{lagrangian} giving:
\begin{equation}
\begin{aligned}
& \underset{u_L} \min \underset{p_L(x) \leq D_L(x)} \max \int_\Omega u_L(x)p_L(x)dx \\
=& \underset{u_L(x) \geq 0} \min  \int_\Omega u_L(x)D_L(x)dx
\end{aligned}
\end{equation}
when $u_L(x) \geq 0$. (If $u_L(x) < 0$, the function can be arbitrarily maximized by $p_L(x) \to -\infty$.) Working through $\mathbb{O}$, every label, $L$, where $L.C \neq \emptyset$ and $L.P \neq \emptyset$ can be isolated in \eqref{lagrangian} as:
\begin{equation}
\underset{u} \min \underset{p_L(x)} \max \left( \int_\Omega u_L(x)p_L(x)dx - \sum_{\forall L' \in L.C} \int_\Omega w_{(L,L')} u_{L'}(x)p_L(x)dx \right) = 0
\end{equation}
at the saddle point defined by $u_L(x) =  \sum_{\forall L' \in L.C} w_{(L,L')} u_{L'}(x)$. Lastly, the source flow, $p_S$, can be isolated in a similar manner, that is:
\begin{equation}
\underset{u} \min \underset{p_S(x)} \max \left( \int_\Omega p_S(x)dx - \sum_{\forall L' \in S.C} \int_\Omega w_{(S,L')} u_{L'}(x)p_S(x)dx \right) = 0
\end{equation}
at the saddle point defined by $1 =  \sum_{\forall L' \in S.C} w_{(S,L')} u_{L'}(x)$. These constraints combined yield the labeling constraints in the original formulation. The maximization of the spatial flow functions can be expressed in a well-studied form \cite{giusti_minimal_1984} as:
\begin{equation} \label{eq:regularSpatialFlow}
\begin{aligned}
\int_\Omega u_L(x) \dvg q_L(x)dx
& = \int_\Omega \left( \dvg (u_L(x)q_L(x)) - q(x) \cdot \nabla u_L(x) \right) dx \\
& = \int_\Omega \dvg (u_L(x)q_L(x)) dx  - \int_\Omega  q_L(x) \cdot \nabla u_L(x) dx \\
& = \oint_{\delta\Omega} u_L(x)q_L(x) \cdot d\mathbf{s}  - \int_\Omega  q_L(x) \cdot \nabla u_L(x) dx \\
& = - \int_\Omega  q_L(x) \cdot \nabla u_L(x) dx \\
\underset{|q_L| \leq \alpha_L S_L(x)} \max \int_\Omega u_L(x) \dvg q_L(x)dx
& = \underset{|q_L| \leq \alpha_L S_L(x)} \max  - \int_\Omega  q_L(x) \cdot \nabla u_L(x) dx \\
& =  - \int_\Omega  \left( -\frac{\alpha_L S_L(x)}{|\nabla u_L(x)|} \nabla u_L(x) \right) \cdot \nabla u_L(x) dx \\
& = \int_\Omega \alpha_L S_L(x) |\nabla u_L(x)| dx \\
\underset{u_L \geq 0} \min \underset{|q_L| \leq \alpha_L S_L(x)} \max \int_\Omega u_L(x) \dvg q_L(x)dx 
& = \underset{u_L \geq 0} \min \int_\Omega \alpha_L S_L(x) |\nabla u_L(x)| dx \text{ .}
\end{aligned}
\end{equation}
The above implies that we can express the saddle point of equation \eqref{lagrangian} as the original energy functional, \eqref{dagmf_form}, and therefore, finding the saddle point of \eqref{lagrangian} is equivalent to solving the DAGMF segmentation problem.

\newpage
\section{Solution to Primal-Dual Formulation}
To address the optimization problem, we can find this saddle point by augmenting the Lagrangian function \cite{bertsekas_nonlinear_1999}:
\begin{equation}
\label{augmented}
\begin{aligned}
\underset{u} \min \underset{p,q} \max & \left( \int_\Omega p_S(x) dx + \sum_{\forall L \neq S} \int_\Omega u_L(x) G_L(x) dx - \frac{c}{2} \sum_{\forall L \neq S} \int_\Omega G_L(x)^2 dx \right)\\
p_L(x) & \leq D_L(x), \text{ } \forall L (L.C = \emptyset) \\
|q_L(x)| & \leq \alpha_L S_L(x) \text{ } L \neq S
\end{aligned}
\end{equation}
where $c$ is a positive penalty parameter encouraging faster convergence to solutions that fulfill the optimization constraints. Using this formula, we can iteratively maximize each component. The solution steps are:
{
\begin{enumerate}
\setlength{\belowdisplayskip}{2pt} \setlength{\belowdisplayshortskip}{2pt}
\setlength{\abovedisplayskip}{2pt} \setlength{\abovedisplayshortskip}{2pt}
\item Maximize \eqref{augmented} over $q_L$ at each vertex by:
\begin{equation*}
q_L(x) \gets \operatorname{Proj}_{|q_L(x)| \leq \alpha_L S_L(x) } \left( q_L(x) + \tau \nabla \left( \dvg q_L(x) + p_L(x) - \sum_{L' \in L.P} w_{(L',L)} p_{L'}(x) - u_L(x)/c \right) \right) 
\end{equation*}
which is a Chambolle's projection iteration. \cite{chambolle_algorithm_2004} $\tau$ is a small positive gradient descent parameter.
\item Maximize \eqref{augmented} the out-flow to the sink, $p_L$ where $L.C = \emptyset$,analytically by:
\begin{equation*}
p_L(x) \gets \min\lbrace D_L(x), \sum_{L' \in L.P} w_{(L',L)} p_{L'}(x) - \dvg q_L(x) + u_L(x)/c \rbrace
\end{equation*}
\item Maximize \eqref{augmented} the flow between vertices, $p_L$ where $L.C \neq \emptyset$ and $L.P \neq \emptyset$, analytically by:
\begin{equation*}
\begin{aligned}
p_L(x) \gets \frac{1}{1+\sum_{L'\ in L.C}w_{(L,L')}^2} & \left( \sum_{L' \in L.P} w_{(L',L)} p_{L'}(x) - \dvg q_L(x) + u_L(x)/c \right.\\
& \left. + \sum_{\forall L' \in L.C}w_{(L,L')}\left( p_{L'}(x) + \dvg q_{L'}(x) + \sum_{L'' \in L'.P/L}w_{(L'',L')} p_{L''}(x) - u_{L'}(x)/c \right) \right)
\end{aligned}
\end{equation*}
\item Maximize \eqref{augmented} over the source flow, $p_S$, analytically by:
\begin{equation*}
\begin{aligned}
p_S(x) \gets \frac{1}{1+\sum_{L'\ in S.C}w_{(S,L')}^2} & \left( 1/c + \right. \\
& \left. \sum_{\forall L' \in S.C} w_{(S,L')} \left( p_{L'}(x) + \dvg q_{L'}(x) + \sum_{L'' \in L'.P/S} w_{L'',L'}p_{L''}(x) - u_{L'}(x)/c \right) \right)
 \end{aligned}
\end{equation*}
\item Minimize \eqref{augmented} over $u_L$ at each vertex analytically by:
\begin{equation*}
u_L(x) \gets u_L(x) - c \left( \dvg q_L(x) - \sum_{L' \in L.P} w_{(L',L)} p_{L'}(x) + p_L(x) \right)
\end{equation*}
\end{enumerate}
}

Note that within each step there exists a large amount of inherent parallelism, that is, each voxel $x$ can be accounted for completely independently of all other voxels in steps 2-5, and with only a dependence on a local neighbourhood in step 1. This inherent parallelism allows for GPGPU acceleration of each step with relative ease.

\newpage
\subsection{Directed Acyclic Graphical Max-Flow Algorithm}
To improve the convergence rate, we perform an initialization step that ensures optimality for the zero-smoothness condition. This is achieved by initializing the system with optimal flows and multipliers under the assumption that all spatial flows are zero. To ensure faster convergence, we order the tasks using a topological sort over the graph. In this ordering, each child label occurs only after all of its parents. The inverse ordering $\mathbb{O}^{-1}$ is the opposite. This is equivalent to the bottom-up approach used by Baxter et al. \cite{baxterGHMF2014} over hierarchies. In addition, each label is equipped with two `working' buffers, $\rho_L(x)$ and $\sigma_L(x)$,  for the purposes of accumulation.

\begin{algorithm}[!h]
Topological sort ($\mathbb{V} \cup \{ S \} $, $\{ (L_1,L_2) |  L_2 \in L_1.P \}$) into ordering $\mathbb{O}$ with reverse ordering $\mathbb{O}^{-1}$\;
InitializeSolution() \;
\While{not converged} {
  UpdateFlows() \;
  \For{ $\forall L$ } {
    $\forall x, u_L(x) \gets u_L(x) - c \left( \dvg q_L(x) - \rho_L(x) + p_L(x) \right)$ \;
  }
}
\label{alg:solverSeq}
\end{algorithm}

which  makes use of the following function definitions:

\begin{algorithm}[!h]
{ \bf UpdateFlows() } \\
\For{ $\forall L \neq S$ } {
  $\forall x, q_L(x) \gets \operatorname{Proj}_{|q_L(x)| \leq S_L(x) } \left( q_L + \tau \nabla \left( \dvg q_L(x) + p_L(x) - \rho_{L}(x) - u_L(x)/c \right) \right)$ \;
  }
Clear $\rho_L(x)$ for all labels \;
\For{ each $L$ in order $\mathbb{O}$ }{
  \For{ each $L' \in L.C$ }{
    $\forall x, \rho_{L'}(x) \gets \rho_{L'}(x) + w_{(L,L')} p_L(x)$ \;
  }
  \uIf{$L.C \neq \emptyset$ and $L.P \neq \emptyset$ }{
  	$\forall x, \sigma_L(x) \gets \rho(x) - \dvg q_L(x) + u_L(x)/c$ \;
  }
  \ElseIf{$L = S$}{
  	$\forall x, \sigma_S(x) \gets 1/c$ \;
  }
}
\For{ each $L$ in order $\mathbb{O}^{-1}$ }{
  \uIf{ $L.C = \emptyset$ }{
    $\forall x, p_L(x) \gets \min\lbrace D_L(x), \rho_L(x) - \dvg q_L(x) + u_L(x)/c \rbrace$ \;
    \For{ $L' \in L.P$}{
      $\forall x, \sigma_{L'}(x) \gets \sigma_{L'}(x) + w_{(L',L)}\left( \dvg q_{L'}(x) + p_{L'}(x) -\rho_{L'}(x) + w_{L',L} p_{L}(x) \right) $\;
    }
  }
  \uElseIf{ $L = S$ }{
    $\forall x, p_S(x) \gets \frac{1}{\sum_{L' \in S.C}w_{(S,L')}^2}\sigma_S(x) $ \;
  }
  \Else{
    $\forall x, p_L(x) \gets \frac{1}{1+\sum_{L' \in L.C}w_{(L,L')}^2} \sigma_L(x)$ \;
    \For{ $L' \in L.P$}{
      $\forall x, \sigma_{L'}(x) \gets \sigma_{L'}(x) + w_{(L',L)}\left( \dvg q_{L'}(x) + p_{L'}(x) -\rho_{L'}(x) + w_{L',L} p_{L}(x) \right) $\;
    }
  }
}
\end{algorithm}

\begin{algorithm}[!h]
{ \bf InitializeSolution() } \\
Clear $u_L(x),q_L(x)$ for all labels\;
\For{ each $L$ in order $\mathbb{O}^{-1}$ }{
  $\forall x, p_L(x) \gets \underset{L'.C = \emptyset}\min D_{L'}(x)$ \;
  $\forall x, \rho_L(x) \gets \underset{L'.C = \emptyset}\min D_{L'}(x)$ \;
  \If{ $L.C = \emptyset$  }{
    \eIf{ $L \in \underset{L'.C = \emptyset}\argmin D_{L'}(x) $ } {
      $\forall x, u_L(x) \gets 1 / | \underset{L'.C = \emptyset}\argmin D_{L'}(x) | $ \;
    }{
      $\forall x, u_L(x) \gets 0$ \;
    }
  }
  \For{ each $L' \in L.P/\{ S \}$  }{
    $\forall x, u_{L'}(x) \gets u_{L'}(x) + w_{(L',L)} u_{L}(x)$ \;
  }
}
\end{algorithm}
\newpage

For the sake of conciseness, the `{\bf for} \textit{$\forall x$} {\bf do} ' loops surrounding each assignment operation have been replaced with the prefix $\forall x$ in both the algorithm and the function definitions.

\section{Regularization of Arbitrary Super-Objects}\label{sec:arbitraryReg}
As demonstrated in the introduction, continuous max-flow segmentation models have been progressing towards more and more general regularization structures. In terms of a discrete analogue, the most general structure possible would be:
\begin{equation}
\begin{aligned}
\underset{ \{\Omega_L\} }{\min} \text{ }  E = \sum\limits_{\forall L} \left( \int_{\Omega_L} D_L(x)dx+ \int_{\delta \Omega_L} S_{L}(x) dx \right)
\end{aligned}
\end{equation}
in which $L$ could refer to either an end-label in $\mathbb{L}$ or a subset $H \subset \mathbb{L}$ with $\Omega_L = \bigcup_{L' \in H} \Omega_{L'}$. We refer to this problem as having \textit{arbitrary super-object regularization} in that it incorporates all possible regularization while maintaining the discrete analogue for intermediate labels. It is easy to see it as a generalization of Potts\cite{potts_generalized_1952}, Ishikawa\cite{ishikawa_exact_2003}, and General Hierarchical\cite{baxterGHMF2014} models. This section aims to illustrate this as a subclass of that represented by DAGMF.

\begin{figure}[h]
\centering
\includegraphics[width = \textwidth]{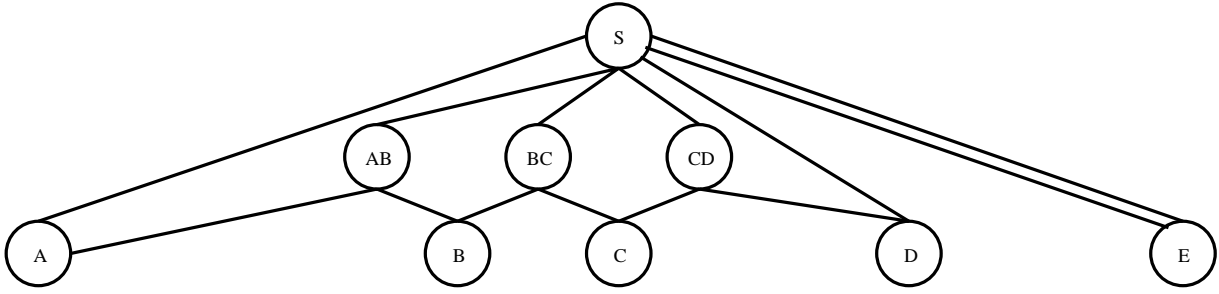}
\caption{DAG for segmentation into labels $\mathbb{L}=\{A,B,C,D,E\}$ in which label groups $\mathbb{G}=\{AB,BC,CD\}$ are regularized. Note that this would be impossible in a hierarchical model since the regularization groups conflict with each other.}
\label{fig:construction1}
\end{figure}

To show how arbitrary super-object regularization can be implemented with DAGMF, we must consider the construction of a DAG with associated transformations on smoothness parameters. First, let us add a vertex to the graph for the source node, $S$, and one for each end-label in $\mathbb{L}$. To do so, let us consider $\mathbb{G} \subset 2^\mathbb{L}$ to be the set of super-objects represented in the problem, not including end-labels. We will associate a vertex in the DAG to each element, $G$, in this set. Each of these vertices has $S$ as their sole parent, and their only children will be vertices corresponding to the end-labels in $G$. Lastly, we must ensure that all end-label vertices have the same number of parents, so we can add edges between $S$ and each vertex (allowing multiplicity) until they do. (Note that each end-label will have $r \leq |\mathbb{G}|$ parents.) An example of this is given in Figure \ref{fig:construction1}.

\begin{figure}[h]
\centering
\includegraphics[width = \textwidth]{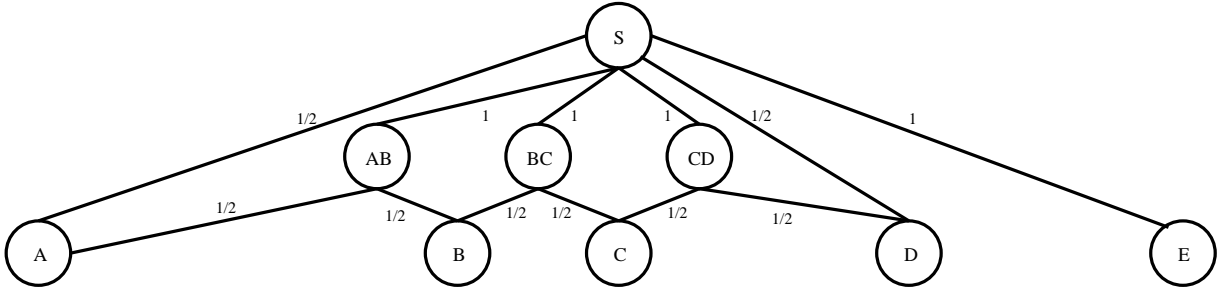}
\caption{DAG from Figure \ref{fig:construction1} with weights explicitly recorded rather than multiplicities.}
\label{fig:construction2}
\end{figure}

Before the algorithm can be used, we must determine edge weights and eliminate multiplicities in the edges. We do this by associated the unnormalized edge weight with the multiplicity of the edge, followed by a normalization step to ensure the weights are valid. The corresponding example is given in Figure \ref{fig:construction2}.

Returning to the general case, if we consider the equation associated with this graph, we will find it is:
\begin{equation}
\begin{aligned}
& \underset{u}{\text{minimize}}
& & E(u) = \sum\limits_{\forall L \in \mathbb{L}} \int_\Omega \left( D_L(x)u_L(x)+S_{L}(x)|\nabla u_L(x)| \right) dx +
 \sum\limits_{\forall L \in \mathbb{G}} \int_\Omega S_{G}(x)|\nabla u_G(x)| dx \\
& \text{subject to} & & \forall L (u_L(x) \geq 0) \\
& & &  \forall G \in \mathbb{G} \left(\sum\limits_{ L' \in G.C }\frac{1}{r}u_{L'}(x) = u_G(x)\right) \\
& & &   u_S(x)= \sum\limits_{ L' \in S.C } w_{(S,L')} u_{L'}(x) =1
\end{aligned}
\end{equation}
which is equivalent to the following by multiplying the labelling functions for $G \in \mathbb{G}$ by $r$
\begin{equation}
\begin{aligned}
& \underset{u}{\text{minimize}}
& & E(u) = \sum\limits_{\forall L \in \mathbb{L}} \int_\Omega \left( D_L(x)u_L(x)+S_{L}(x)|\nabla u_L(x)| \right) dx +
 \sum\limits_{\forall L \in \mathbb{G}} \int_\Omega \frac{1}{r}S_{G}(x)|\nabla u_G(x)| dx \\
& \text{subject to} & & \forall L (u_L(x) \geq 0) \\
& & &  \forall L \in \mathbb{G} \left(\sum\limits_{ L' \in L.C } u_{L'}(x) = u_L(x)\right) \\
& & &   u_S(x) = \sum_{L\in\mathbb{L}}u_L(x)=1 \text{ .}
\end{aligned}
\end{equation}
This indicates that the intended regularization field for $G \in \mathbb{G}$ must be multiplied by $r$ for correct scaling, and for the union operator to be a discrete analogue for addition (since the multiplicand is removed in the constraint equations).

\section{Conclusions}
In this paper, we present an algorithm for addressing max-flow segmentation problems where the underlying architecture is a directed acyclic graph. Using this architecture, we can achieve arbitrary super-object regularization, making it more general version of Generalized Hierarchical Max-Flow\cite{baxterGHMF2014} and thus both Potts and Ishikawa models. Such a regularization mechanism allows for the definition of more flexible part/whole relationships, where any individual part could belong to more than one whole, a capability not present in any earlier extendable model. These also represent the most general form of ordering relationships possible while maintaining an analogous purely discrete problem.

This solver has been implemented using the NVIDIA Compute Unified Device Architecture (CUDA) taking advantage of the inherent parallelism within each optimization step. Additional concurrency between steps has been exploited to improve computational speed and allow for multiple graphics cards to be used simultaneously on a single segmentation problem.

%%%%%%%%%%%%%%%%%%%%%%%%%%%%%%%%%%%%%%%%%%%%%%%%%%%%%%%%%%%%%
\acknowledgments %>>>> equivalent to \section*{ACKNOWLEDGMENTS}
The authors would like to acknowledge Dr.\ Elvis Chen, Kamyar Abhari, and Jonathan McLeod for their invaluable discussion, editing, and technical support.

%%%%%%%%%%%%%%%%%%%%%%%%%%%%%%%%%%%%%%%%%%%%%%%%%%%%%%%%%%%%%
%%%%% References %%%%%

\bibliographystyle{spiebib}   %>>>> makes bibtex use spiebib.bst
\bibliography{TechReportDAGMF} 

\begin{thebibliography}{10}

\bibitem{boykov_interactive_2001}
Y.~Y. Boykov and M.-P. Jolly, ``Interactive graph cuts for optimal boundary \&
  region segmentation of objects in {ND} images,'' in {\em Computer Vision,
  2001. {ICCV} 2001. Proceedings. Eighth {IEEE} International Conference on},
  {\bf 1}, pp.~105–--112, 2001.

\bibitem{boykov_fast_2001}
Y.~Boykov, O.~Veksler, and R.~Zabih, ``Fast approximate energy minimization via
  graph cuts,'' {\em {IEEE} Transactions on Pattern Analysis and Machine
  Intelligence}~{\bf 23}(11), pp.~1222--1239, 2001.

\bibitem{boykov_experimental_2004}
Y.~Boykov and V.~Kolmogorov, ``An experimental comparison of min-cut/max-flow
  algorithms for energy minimization in vision,'' {\em Pattern Analysis and
  Machine Intelligence, {IEEE} Transactions on}~{\bf 26}(9), pp.~1124–--1137,
  2004.

\bibitem{boykov_markov_1998}
Y.~Boykov, O.~Veksler, and R.~Zabih, ``Markov random fields with efficient
  approximations,'' in {\em Computer vision and pattern recognition, 1998.
  Proceedings. 1998 {IEEE} computer society conference on},  pp.~648--–655,
  1998.

\bibitem{li_markov_2009}
S.~Z. Li, {\em Markov Random Field Modeling in Image Analysis}, Springer, Jan.
  2009.

\bibitem{yuan_study_2010}
J.~Yuan, E.~Bae, and X.-C. Tai, ``A study on continuous max-flow and min-cut
  approaches,'' in {\em Computer Vision and Pattern Recognition ({CVPR)}, 2010
  {IEEE} Conference on},  pp.~2217--–2224, 2010.

\bibitem{ishikawa_exact_2003}
H.~Ishikawa, ``Exact optimization for markov random fields with convex
  priors,'' {\em {IEEE} Transactions on Pattern Analysis and Machine
  Intelligence}~{\bf 25}(10), pp.~1333--1336, 2003.

\bibitem{bae_fast_2011}
E.~Bae, J.~Yuan, X.-C. Tai, and Y.~Boykov, ``A fast continuous max-flow
  approach to non-convex multi--labeling problems,'' 2011.

\bibitem{delong2009globally}
A.~Delong and Y.~Boykov, ``Globally optimal segmentation of multi-region
  objects,'' in {\em Computer Vision, 2009 IEEE 12th International Conference
  on},  pp.~285--292, IEEE, 2009.

\bibitem{baxterGHMF2014}
J.~S. Baxter, M.~Rajchl, J.~Yuan, and T.~M. Peters, ``A continuous max-flow
  approach to general hierarchical multi-labelling problems,'' {\em arXiv
  preprint arXiv:1404.0336} , 2014.

\bibitem{rajchl_interactive_2014}
M.~Rajchl, J.~Yuan, J.~White, E.~Ukwatta, J.~Stirrat, C.~Nambakhsh, F.~Li, and
  T.~Peters, ``Interactive hierarchical max-flow segmentation of scar tissue
  from late-enhancement cardiac {MR} images,'' {\em {IEEE} Transactions on
  Medical Imaging} , 2014.

\bibitem{yuan_efficient_2012}
J.~Yuan, W.~Qiu, E.~Ukwatta, M.~Rajchl, Y.~Sun, and A.~Fenster, ``An efficient
  convex optimization approach to {3D} prostate {MRI} segmentation with generic
  star shape prior,'' {\em Prostate {MR} Image Segmentation Challenge,
  {MICCAI}} , 2012.

\bibitem{potts_generalized_1952}
R.~B. Potts, ``Some generalized order-disorder transformations,'' in {\em
  Proceedings of the Cambridge Philosophical Society},   {\bf 48},
  pp.~106--–109, 1952.

\bibitem{yuan_continuous_2010}
J.~Yuan, E.~Bae, X.-C. Tai, and Y.~Boykov, ``A continuous max-flow approach to
  potts model,'' in {\em Computer Vision – {ECCV} 2010},  K.~Daniilidis,
  P.~Maragos, and N.~Paragios, eds., {\em Lecture Notes in Computer Science},
  pp.~379--392, Springer Berlin Heidelberg, Jan. 2010.

\bibitem{bae_global_2011}
E.~Bae, J.~Yuan, and X.-C. Tai, ``Global minimization for continuous multiphase
  partitioning problems using a dual approach,'' {\em International journal of
  computer vision}~{\bf 92}(1), pp.~112–--129, 2011.

\bibitem{kolmogorov_computing_2001}
V.~Kolmogorov and R.~Zabih, ``Computing visual correspondence with occlusions
  using graph cuts,'' in {\em Computer Vision, 2001. {ICCV} 2001. Proceedings.
  Eighth {IEEE} International Conference on},   {\bf 2}, pp.~508–--515, 2001.

\bibitem{ekeland_convex_1976}
I.~Ekeland and R.~Temam, ``Convex analysis and variational problems,'' 1976.

\bibitem{giusti_minimal_1984}
E.~Giusti, {\em Minimal surfaces and functions of bounded variation}, vol.~80,
  Birkhauser, 1984.

\bibitem{bertsekas_nonlinear_1999}
D.~P. Bertsekas, ``Nonlinear programming,'' 1999.

\bibitem{chambolle_algorithm_2004}
A.~Chambolle, ``An algorithm for total variation minimization and
  applications,'' {\em Journal of Mathematical imaging and vision}~{\bf
  20}(1-2), pp.~89–--97, 2004.

\end{thebibliography}

\end{document}